\documentclass[10pt,twocolumn,letterpaper]{article}

\usepackage[pagenumbers]{cvpr} 

\usepackage{graphicx}
\usepackage{amsmath}
\usepackage{amssymb}
\usepackage{booktabs}
\usepackage{multirow} 
\usepackage{comment}
\usepackage{grffile}

\usepackage{hyperref}
\hypersetup{pagebackref,breaklinks,colorlinks}

\usepackage[capitalize]{cleveref}
\crefname{section}{Sec.}{Secs.}
\Crefname{section}{Section}{Sections}
\Crefname{table}{Table}{Tables}
\crefname{table}{Tab.}{Tabs.}

\def\cvprPaperID{3167} 
\def\confName{CVPR}
\def\confYear{2023}

\begin{document}

\title{Compressing Volumetric Radiance Fields to 1 MB}

\author{Lingzhi Li$^*$, Zhen Shen$^*$, Zhongshu Wang, Li Shen, Liefeng Bo\\
Alibaba Group\\
Beijing, China\\
{\tt\small \{llz273714,zackary.sz, zhongshu.wzs, jinyan.sl, liefeng.bo\}@alibaba-inc.com}
}
\maketitle

\begin{abstract}
Approximating radiance fields with volumetric grids is one of promising directions for improving NeRF, represented by methods like Plenoxels and DVGO, which achieve super-fast training convergence and real-time rendering. However, these methods typically require a tremendous storage overhead, costing up to hundreds of megabytes of disk space and runtime memory for a single scene. We address this issue in this paper by introducing a simple yet effective framework, called vector quantized radiance fields (VQRF), for compressing these volume-grid-based radiance fields. We first present a robust and adaptive metric for estimating redundancy in grid models and performing voxel pruning by better exploring intermediate outputs of volumetric rendering. A trainable vector quantization is further proposed to improve the compactness of grid models. In combination with an efficient joint tuning strategy and post-processing, our method can achieve a compression ratio of 100$\times$ by reducing the overall model size to 1 MB with negligible loss on visual quality. Extensive experiments demonstrate that the proposed framework is capable of achieving unrivaled performance and well generalization across multiple methods with distinct volumetric structures, facilitating the wide use of volumetric radiance fields methods in real-world applications. Code Available at  \url{https://github.com/AlgoHunt/VQRF}

{\let\thefootnote\relax\footnotetext{$^{*}$denote equal contribution}}

\end{abstract}


\section{Introduction}
\begin{figure}[t]  
    \centering
    \includegraphics[width=1.0\linewidth]{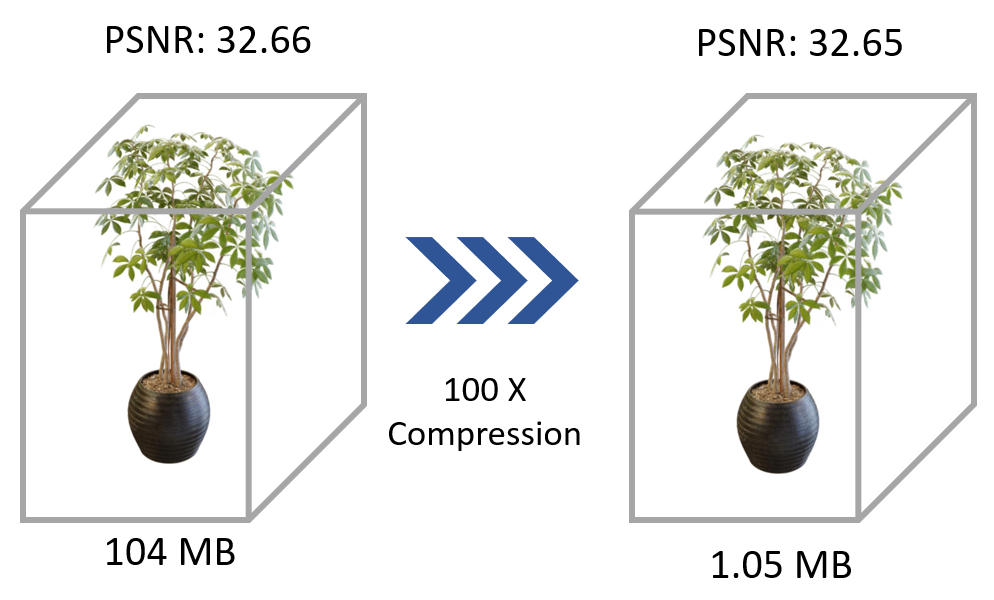}
    \caption{Our compression pipeline realizes 100X compression rate while maintaining the rendering quality of the original volumetric model.} 
    \label{fig:teaser}   
\end{figure}

Novel view synthesis aims to realize photo-realistic rendering for a 3D scene at unobserved viewpoints, given a set of images recorded from multiple views with known camera poses. The topic has growing importance because of its potential use in a wide range of Virtual Reality and Augmented Reality applications. Neural radiance fields (NeRF) \cite{NeRF} have demonstrated compelling ability on this topic by modelling and rendering 3D scenes effectively through the use of deep neural networks, which are learned to map each 3D location given a viewing direction to its corresponding view-dependent color and volume density according to volumetric rendering techniques \cite{OpticalModel}. The rendering process relies on sampling a huge number of points and feeding them through a cumbersome network, incurring considerable computational overhead during training and inference. Recent progress following radiance fields reconstruction shows that integrating voxel-based structures \cite{SparseVoxel} into the learning of representations can significantly boost training and inference efficiency. These volumetric radiance fields methods typically store features on voxels and retrieve  sampling points (including color features and volume densities) by performing efficient trilinear interpolation without neural network \cite{yu2021plenoxels} or only equipped with a lightweight neural network \cite{ChengSun2022DirectVG} instead of cumbersome networks. However, the use of volumetric representations inevitably introduces considerable storage cost, e.g., costing over one hundred megabytes to represent a scene (shown in the Fig.~\ref{fig:teaser}, which is prohibitive in real-world applications. 

\begin{figure*}[t]  
    \centering
    \includegraphics[width=1.0\linewidth]{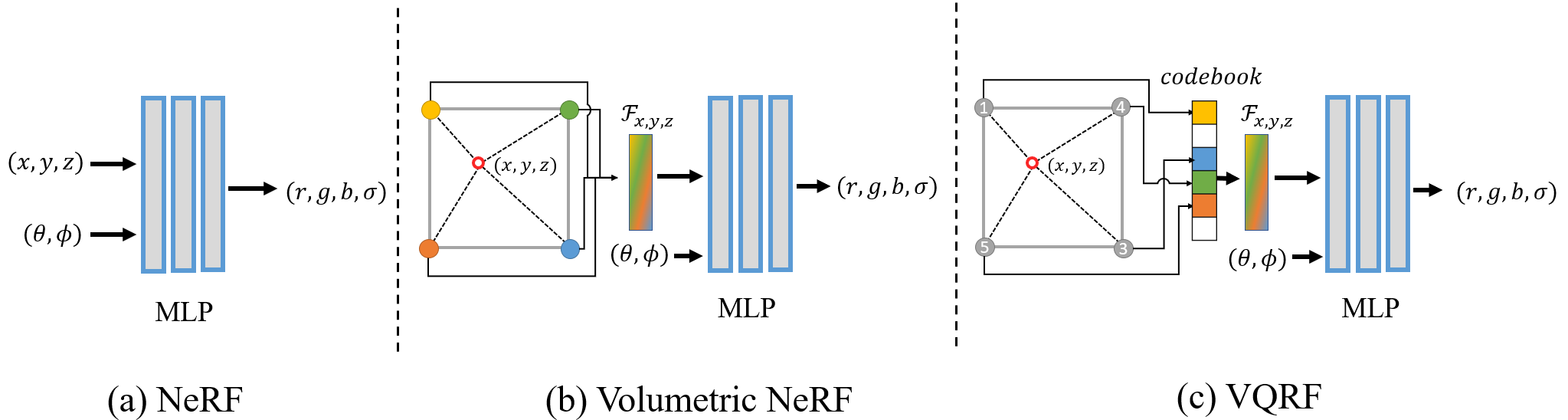}
    \caption{(a) NeRF learns a mapping from 3d coordinate $(x,y,z)$ and viewing direction$(\theta,\phi)$ to color and density $(r,g,b,\sigma)$. (b) Volumetric NeRF optimizes  a volumetric grid and estimating color feature $\mathcal{F}_{x,y,z}$ for sampling point via tri-linear interpolation. (c) Our VQRF stores $k$-bits mapping index per voxel. The index point to the actual feature inside a codebook consisting of  $2^k$ codes. } 
    \label{fig:three type of rf}   
\end{figure*}
In this paper, we aim to counteract the storage issue of representations induced by using voxel grids meanwhile retaining rendering quality. In order to better understand the characteristic of grid models, we estimated the distribution of voxel importance scores (shown in Fig.~\ref{fig:cdf_qq}) and observed that only 10\% voxels contribute over 99\% importance scores of a grid model, indicating that large redundancy exists in the model. Inspired by traditional techniques of deep network compression \cite{han2015deep}, we present an effective and efficient framework for compressing volumetric radiance fields, allowing about $100\times$ storage reduction over original grid models, with competitive rendering quality.

\begin{figure*}[t]  
    \centering
    \includegraphics[width=1.0\linewidth]{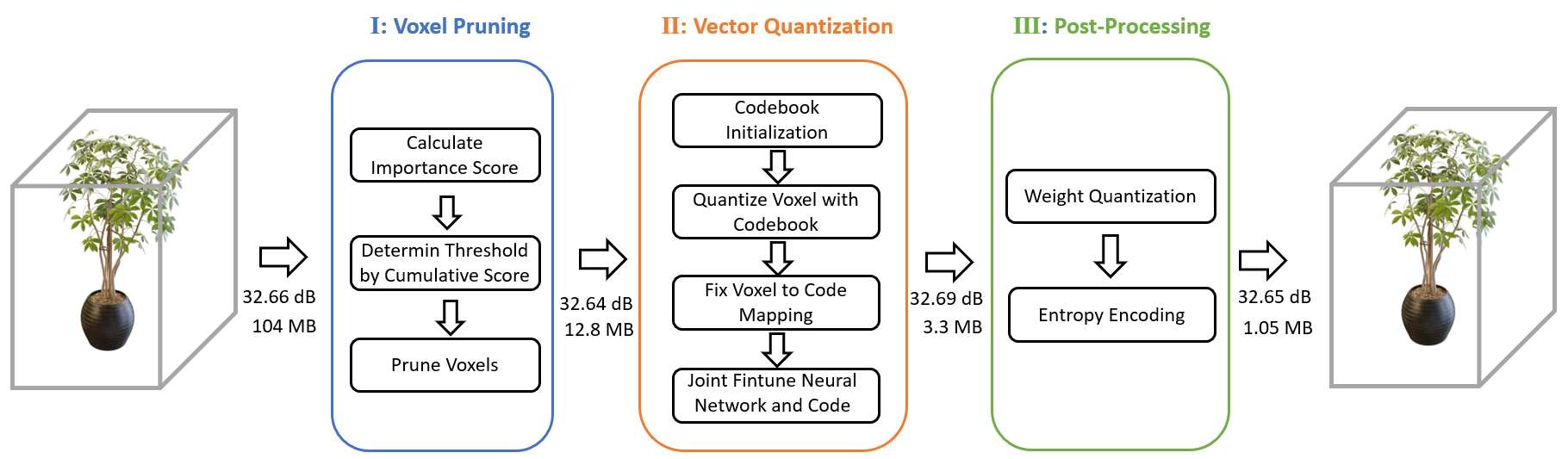}
    \caption{\textbf{Pipeline overview.} We design a three stage compression pipeline, given a DVGO trained on \textit{"ficus"} scene for example, voxel prune reduce model size by 8$\times$, vector quantization further improve compression ratio to around 33 $\times$, and we achieve a total of 100$\times$ compression with addtional post-processing.} 
    \label{fig:pipline}   
\end{figure*}

The illustration of the framework is shown in Fig.~\ref{fig:pipline}. The proposed framework is quite general rather than restricted to certain architecture. The overall framework is comprised of three steps, i.e., voxel pruning, vector quantization and post processing. Voxel pruning is used to omit the least important voxels which dominate model size while contributing little to the final rendering. We introduce an adaptive strategy for pruning threshold selection with the aid of a cumulative score rate metric, enabling the pruning strategy general across different scenes or base models. In order to further reduce model size, we propose to encode important voxel features into a compact codebook by developing importance-aware vector quantization with an efficient optimization strategy, where a joint tuning mechanism encourages the compressed models approaching to rendering quality of original models. We finally perform a simple post-processing step to obtain a model with quite small storage cost. For example, as shown in Fig.~\ref{fig:teaser}, the original model a storage cost of 104 MB and PSNR $32.66$ can be compressed into the model costing 1.05 MB with a negligible visual quality loss (PSNR $32.65$). We conduct extensive experiments and empirical studies to validate the proposed compression framework, showing the effectiveness and generalization of the proposed compression pipeline on a wide range of volumetric methods and varying scenarios. 

\section{Related Work}
\label{sec:related_work}

\subsection{Neural Radiance Fields} 
Neural radiance fields\cite{NeRF} are a recently emerging technique. It provides an effective scene representation to enable a high-fidelity novel-view synthesis with only multi-view input. Volumetric methods or explicit methods \cite{yu2021plenoxels,yu2021plenoctrees,TensoRF,ChengSun2022DirectVG} are one of the promising directions for NeRF optimization, especially in training efficiency and rendering acceleration. 

SNeRG \cite{SNeRG} convert NeRF to a carefully designed grid to achieve inference speedup. \cite{yu2021plenoctrees}  take advantage of traditional octree data structure. 
Plenoxels\cite{yu2021plenoxels} directly optimize a sparse spherical harmonic grid to model view-dependent effects without a neural network. 
DVGO\cite{ChengSun2022DirectVG} uses two voxel grid to represent both color features and density.
PlenOctrees\cite{yu2021plenoctrees} make use of octree, InstantNGP\cite{InstantNGP} adopted a multi-scale hash table framework, and TensorRF\cite{TensoRF} utilizes tensor decomposition\cite{TamaraGKolda2009TensorDA} to approximate full grid feature with tri-plane. All those methods, especially the volumetric one, will inevitably enlarge the storage overhead compared to NeRF's pure mlp design.

\subsection{Vector Quantization}
Vector quantization\cite{gray1984vector, gersho2012vector, equitz1989new} is a classical lossy compression technique. It has been widely used in many real-world applications, including image compression\cite{nasrabadi1988image, cosman1993using}, video codec\cite{lee1995motion,sikora1997mpeg} and audio codec \cite{paliwal1993efficient,makhoul1985vector}. The main idea of vector quantization is to cluster a large set of vectors into a smaller set of clusters and present each vector with corresponding cluster centroid. \cite{gong2014compressing} introduce vector quantization to deep neural network compression.\cite{takikawa2022variable} utilize a soft vector quantization to compress neural fields. Vector quantization has also been used in generative model \cite{van2017neural, gu2022vector, razavi2019generating}, but was meant to improve generation quality rather than compression.

\subsection{Model Compression}
 Another most related topic is model compression, which aims to reduce overall storage size while preserving  maximum accuracy of the uncompressed model.
 Most model compression technique fall into four major classes: (1) model pruning\cite{vanhoucke2011improving,gupta2015deep,han2015deep,wen2016learning}, which remove redundant connections or layers from a neural network. (2) weight quantization\cite{jacob2018quantization, han2015deep, krishnamoorthi2018quantizing} reduce model size by convert full precision float number to 8-bits or even lower representation.
 (3) low-rank approximation\cite{rigamonti2013learning,denton2014exploiting,jaderberg2014speeding} decomposes high-rank matrix into a smaller counterpart. (4) knowledge distillation \cite{ba2013deep,hinton2015distilling,chen2017learning,polino2018model} use a well trained large network to guide the training of a compact small network. Those approaches are mostly orthogonal and can be used together, \cite{han2015deep} utilize both model prune, weight quantization and huffman coding to further compress neural network.
 Some volumetric radiance fields have already employed those techniques. PlenOctrees\cite{yu2021plenoctrees},Re:NeRF\cite{deng2022compressing} applies weight quantization. Plenoxels' empty voxel pruning mechanism is similar to weight pruning. CCNeRF\cite{tang2022compressible} and TensoRF\cite{TensoRF} decomposed full-size tensor to its low-rank approximation.
 


\section{Problem Statement}
Neural radiance fields \cite{NeRF} learn a continuous function that maps a 3D point $\mathbf{x} \in \mathbb{R}^3$ and viewing direction $\mathbf{d} \in \mathbb{R}^3$ to the view-dependent color $ \mathbf{c} \in \mathbb{R}^3$ and the volume density $\sigma \in \mathbb{R}$ through the use of a multilayer perceptron (MLP) i.e.
$F_{\Theta}: (\mathbf{x}, \mathbf{d}) \mapsto (\mathbf{c}, \sigma) $. According to the volume rendering technique \cite{OpticalModel}, the pixel color $\widehat{C}(\mathbf{r})$  of a given ray $\mathbf{r} = \mathbf{o} + t\mathbf{d}$ can be estimated by accumulating the color $\mathbf{c}$ and density $\mathbf{d}$ of sampling points along the ray:
\begin{equation}\label{eq:render} 
    \widehat{C}(\mathbf{r}) = \sum_{i=1}^N T_i\cdot \alpha_i \cdot \mathbf{c}_i,
\end{equation}
\begin{equation}\label{eq:alpha_t}
    \alpha_i = 1 - \exp(-\sigma_i \delta_i),\quad T_i = \prod_{j=1}^{i-1} (1 - \alpha_j),
\end{equation}
where $\delta_i$ is the distance between adjacent points. $T_i$ is the accumulated transmittance when reaching the point $i$, and $\alpha_i$
is ray termination probability.

Recently, volumetric radiance fields methods \cite{ChengSun2022DirectVG, yu2021plenoxels} introduce voxel-based structure to facilitate the learning of representations, i.e., optimizing a volumetric grid $\mathcal{V} = \{\mathbf{V}_c, \mathbf{V}_\sigma\}$ and estimating the color features and density for sampling points via tri-linear interpolation.
The methodology has shown significant benefit on training and inference efficiency compared to the methods relying on large neural networks. However, the use of volumetric representations inevitably introduce considerable storage cost, which might limit its usability in real world applications. 

To address the issue, we introduce a simple yet efficient framework to compress volumetric radiance fields with the following operations, voxel pruning, vector quantization, and post-processing with weight quantization and entropy encoding, which will be described in detail in the following sections.


\section{Voxel Pruning}
\label{sec:voxel_prune}
In order to better understand the statistics of volumetric representations, we first compute the importance scores for each voxel in the grid. Formally, according to the volume rendering technique defined in Eqn.\ref{eq:render} and \ref{eq:alpha_t}, we can get the sampling point $\mathbf{x}_i$ which is tri-linearly interpolated with its neighboring voxels $\mathbf{v}_l$ where $\mathbf{v}_l \in \mathcal{N}_i$. 
\begin{equation}
I_i =  T_i \cdot \alpha_{i}, 
\label{eq:point_IS}
\end{equation}
The importance score is assigned to the voxel $\mathbf{v}_l$ proportionally according to its distance to the point $\mathbf{x}_i$. The importance score of the voxel $\mathbf{v}_l$ can then be obtained by accumulating the importance scores of sampling points which contribute to it,

\begin{equation}
I_l =\sum_{\mathbf{x}_i \in\mathcal{N}_l} (1-|\mathbf{v}_l - \mathbf{x}_i|) \cdot I_i 
\label{eq:voxel_IS}
\end{equation}
Where $\mathcal{N}_l$ denotes the set of the sampling points falling within the neighborhood of $v_l$ and $|\mathbf{v}_l - \mathbf{x}_i|\leq 1$. 
In practice, we shoot a batch of cays on the images of the training views and calculate the importance score of each sampling point. The importance score for each voxel can be obtained according to Eqn.~\ref{eq:voxel_IS}.

\begin{figure*}[t]  
    \centering
    \includegraphics[width=0.95\linewidth]{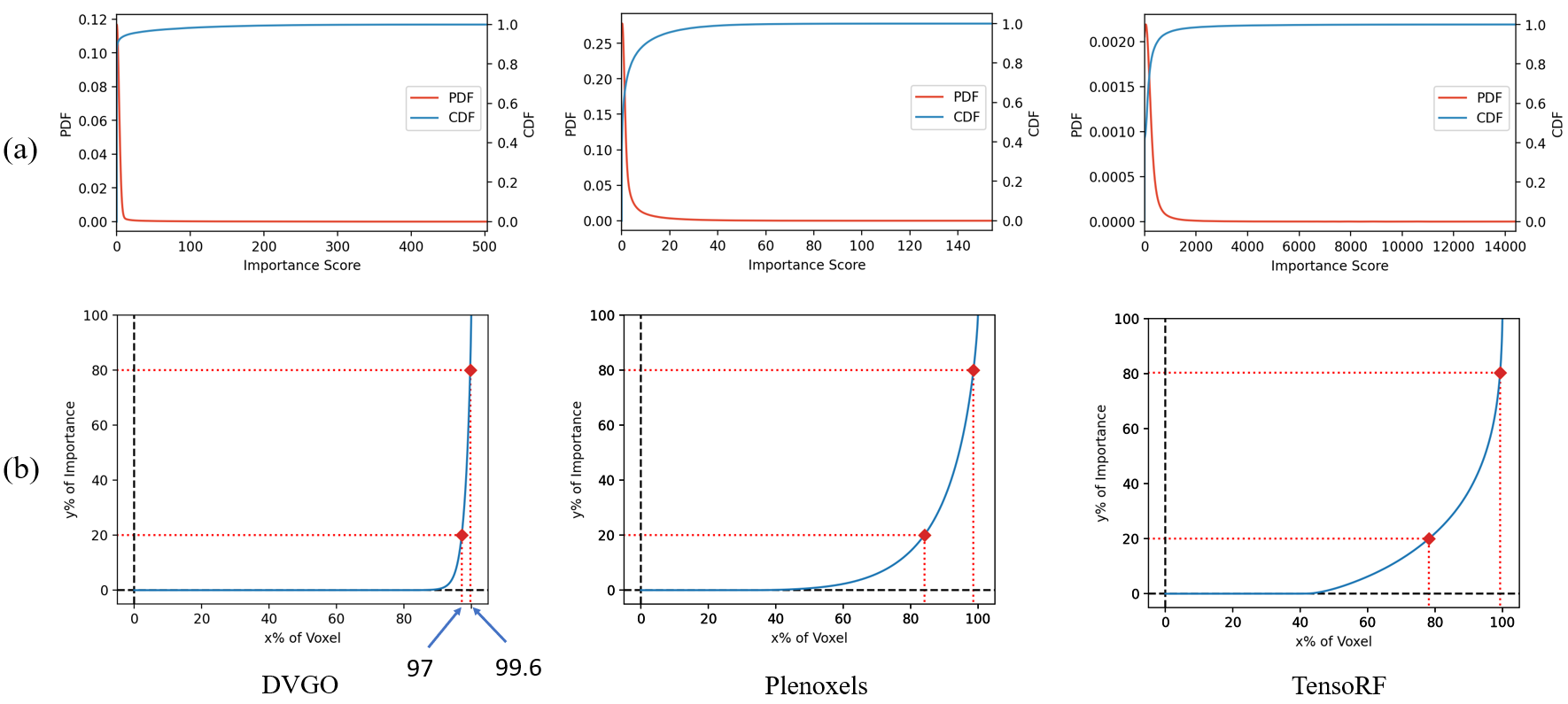}
    \caption{(a) PDF and CDF of importance score. (b) We draw the quantile-quantile curve, which means x\% of least important voxels contributes to y\% percent of total importance. Take DVGO (bottom left curve) for example,  \textbf{97 \%} of least important voxels only contribute 20\% total importance, which equals the importance contributed by the top  \textbf{0.4\%} $(1-99.6\%)$ of voxels. } 
    \label{fig:cdf_qq}   
\end{figure*}

Then we sort the voxels with ascending importance scores and define the cumulative score rate with respect to the parameter $\theta$ as:
\begin{equation}
F(\theta) =   \frac{\sum I_l \cdot \mathbf{1}\{I_l<\theta\}}{\sum I_l},
\label{eq:CDF_IS}
\end{equation}
where $\mathbf{1}{\cdot}$ denotes the binary indicator. The cumulative score rate is proportional to the expectation on the cumulative distribution of voxel importance scores. Take the \textit{Lego} scene in the synthetic-NeRF dataset for example, we depict the curve of the cumulative score rate on the DVGO's model \cite{ChengSun2022DirectVG} in the figure~\ref{fig:cdf_qq}. As shown in the figure, there exist an obviously long-tail phenomenon that more than \textbf{99.9\%} of the importance scores is contributed by the \textbf{10\%} of the voxels. In other words, most of the voxels have minimal effect on rendering results, which indicates large redundancy in the grid model and can be pruned off without scarifying rendering quality. 

We expect the pruning strategy to be fairly general across different scenarios or methods. 
In this regard, we use the quantile function to adaptively select the threshold $\theta_p$ for voxel pruning:

\begin{equation}
\theta_p = F^{-1}(\beta_{p}),
\label{eq:quantile_prune}
\end{equation}
where $\beta_p$ is a hyperparameter that represents the total amount of importance to be pruned. For all the voxels with the importance score lower than $\theta_p$, we directly omit them from the grid model (including densities and color features).

\section{Vector Quantization}

We compress the important voxels to further reduce model size. Compared to density modality, color features typically cost much more storage. In this regard, we adopt a vector quantization strategy to encode voxel color features into a compact codebook, so that the color features of multiple voxels can be substituted by a single code vector. Then the model only needs to store the codebook and the corresponding mapping index from voxels to the codebook instead of storing individual voxels. We will present the training strategy for obtain an effective and compact codebook in the following subsections.

Let us analyze the change of storage cost before and after performing vector quantization. Assume there are $N$ voxels  with the channel dimension $C$, a codebook with the size $K\times C$ is learned in vector quantization, where $K$ is supposed to be extremely smaller than $N$.  Each feature is typically saved in float16 format, so the color features of the orginal voxels would cost $N \times C \times 16$ bits. When applying vector quantization, the storage cost for saving codebook is $16KC$ bits and the index needs $log_2(K)$ bits to present a single voxel point. Using the strategy can achieve the compression rate $r$ on model size as:
\begin{equation}
r = \frac{16NC}{Nlog_2(K) + 16KC}
\label{eq:compression_rate}
\end{equation}

For example, when applying vector quantization with 4096 codes, The upper bound compression rate $r$ would be 16 for DVGO \cite{ChengSun2022DirectVG}, 64 for TensoRF \cite{TensoRF} and 36 for Plenoxels \cite{yu2021plenoxels} at their default setting on the synthetic-NeRF dataset.

\subsection{Codebook Initialization and Update}

We use a weighted clustering strategy for initializing the codebook by the consideration that the voxels with higher importance scores typically have higher impact for rendering. 
Formally, the voxel features $\mathbf{U} = \left \{\mathbf{u}_1, \mathbf{u}_2, ..., \mathbf{u}_N\right \}$ are partitioned into the codebook $\mathbf{B} = \left \{\mathbf{b}_1, \mathbf{b}_2, ..., \mathbf{b}_K\right \}$ where $N \gg  K$, by minimizing the weighted within-cluster sum of squares:
\begin{equation}
\begin{matrix}
\\
\text{arg}\,\text{min}
\\ 
\mathbf{B}
\end{matrix} \sum_{k=1}^K \sum_{\mathbf{v}_j \in \mathcal{R}(\mathbf{b}_k)} \|\mathbf{u}_j - \mathbf{b}_k\|_2^2 \cdot I_j,
\label{eq:WWCS}
\end{equation}
where $\mathbf{u}_j$ and $I_j$ denote the color features and the importance score of $\mathbf{v}_j$, $\mathcal{R}(\mathbf{b}_k)$ denotes the set of voxels assigned to the $k$-the code vector $\mathbf{b}_k$.

In practice, applying the weighted clustering is unpleasantly slow when $n$ and $k$ are large. We introduce an iterative optimization strategy to approximate the procedure. 
Specially, we randomly select a batch of voxels from the grid at each iteration and calculate the euclidean distance between the selected voxel and each code vector in the codebook to determine which code the voxel associates with. The code vector is optimized by weighted accumulating the voxel features belonging to the code by virtue of importance score,
\begin{equation}
 \mathbf{b}_k: = \lambda_d \mathbf{b}_k + (1-\lambda_d) \sum_{\mathbf{v}_j \in \mathcal{R}(\mathbf{b}_k)} I_j \mathbf{u}_j.
\label{eq:centroid_update}
\end{equation}
Here $\lambda_d$ is the decay factor for the moving average of the code vector updating.

\textbf{Code expiration.} 
Using the iterative optimization might encounter inactive code issue, which means some code vectors may only associate with a minimal amount or even zero amount of voxels while some code vectors are likely shared by a large amount informative voxels. It would result in imbalanced assignment distribution which may degrade 
the representation ability of the codebook. To address the issue, we track the capacity of each code vector $\mathbf{b}_k$ by estimating the accumulated importance of the voxels assigned to it during iteration, i.e., $s_k = \sum  I_j \mathbf{1}\{\mathbf{v}_j \in \mathcal{R}(\mathbf{b}_k)\} $. 
Then we rank them in descending order and expire $J$ codes with lowest capacity, which are reinitialized with the features of the top $J$ mostly important voxels in the batch.

\textbf{Which Voxel needs Quantization?}
In order to achieve a good balance between rendering quality loss and compression rate, we save a fraction of mostly important voxels without passing them through the vector quantization, where the fraction rate is determined by the quantile defined in the Eqn.~\ref{eq:quantile_prune} as,
\begin{equation}
\theta_k = Q(\beta_{k}) = F^{-1}(\beta_{k}),
\label{eq:quantile_keep}
\end{equation}
where $\beta_{k}$ denotes the hyperparameter, and $\theta_k$ represents the keeping threshold. The voxels with the important score larger than $\theta_k$ are directly stored, named as non-vector-quantized voxels (\textit{non-VQ voxels}). As shown in Fig.~\ref{fig:cdf_qq}, the statistics reveal that the top 1\% of voxels can contribute over 60\% of the importance, saving a fraction of voxels facilitate rendering quality preservation with  minimal increase on storage, achieving a better trade-off compared to compressing all through VQ.

\subsection{Joint Finetune}
Directly performing vector quantization after voxel pruning can compress a grid model to 5\% of the original size, but it would bring in unacceptable performance loss (from 31.88 to 31.32 in the ablation study in Table \ref{tab:step by step}). In order to improve the effectiveness of the vector quantization strategy, we propose to fine-tune on the features in the voxels grid (as well as MLPs if the original method used) jointly with the VQ optimization. The insight is similar to weight pruning and quantization in deep network compression \cite{han2015deep}, as we expect to tune the compressed grid model to approach the performance of the original model.
Take the compression on DVGO for example. During joint finetuning phase, we \textbf{fix} and save the voxel-to-codebook mapping. Four parts need to be tuned including 1) code vector in the codebook, 2) density grid, 3)  non-VQ voxels and 4) small network originally used in the DVGO.

As the size of voxel gradients are extremely large but sparse, we update each code vector by synchronizing weights across the voxels assigned to it for every $i$ iterations, which can boost training efficiency.

\section{Post-Processing}
We can further reduce model size via the post-processing step, comprised of weight quantization \cite{gholami2021survey} and entropy encoding \cite{mackay2003information}. We use a simple uniform weight quantization on the density voxel and non-vector-quantized feature voxels, without operating on codebook as it is fairly compact. An 8-bit weight quantization casts full-precision floating number to unsigned integers. 

We store two boolean masks to identify which voxel have been pruned, vector quantized or saved unaltered.
The storage for a DVGO model finally comes from saving the following six components, 1) the 2-bits mask, 2) code vectors in the codebook, 
3) mapping indexes between voxels and codebook, 4) 8-bit quantized density grid, 5) 8-bit quantized non-VQ voxels and 6) the small network the method originally used. We encode them with entropy encoding (LZ77\cite{ziv1977universal,ziv1978compression,zlib}) and pack them together to get the final storage cost.

\begin{table*}[t]
\centering
\small
\begin{tabular}{l|ccc|ccc|ccc|ccc}
\toprule
                     & \multicolumn{3}{c|}{Synthetic-NeRF} & \multicolumn{3}{c|}{Synthetic-NSVF} & \multicolumn{3}{c|}{LLFF} & \multicolumn{3}{c}{Tanks\&Temples} \\ 
   Methods          & PSNR       & SSIM      & SIZE      & PSNR       & SSIM      & SIZE      & PSNR   & SSIM  & SIZE    & PSNR       & SSIM      & SIZE      \\

                     & (dB)$\uparrow$  & $\uparrow$ & (MB)$\downarrow$ & (dB)$\uparrow$  & $\uparrow$ & (MB)$\downarrow$ & (dB)$\uparrow$  & $\uparrow$ & (MB)$\downarrow$ & (dB)$\uparrow$  & $\uparrow$ & (MB)$\downarrow$       \\ \midrule \midrule
NeRF\cite{NeRF}      & 31.01      & 0.947     & 5.0      & -           &  -         & -          & 26.50  & 0.811 & 5.0    & 25.78      & 0.864     & 5.0      \\
CCNeRF-CP\cite{tang2022compressible}     & 30.55      & 0.935     & 4.4       & -         &  -        & -         & -      & -     & -       & 27.01      & 0.878     & 4.4       \\
TensoRF-CP\cite{TensoRF}     & 31.56      & 0.949     & 3.9       & 34.48  &  0.971 & 3.9     & -      & -     & -       & 27.59      & 0.897     & 3.9       \\ \midrule \midrule
DVGO\cite{ChengSun2022DirectVG}          & 31.90      & 0.956     &105.9     &34.90     & 0.975      & 119.8    & -      &  -   & -      &  28.29   & 0.910  & 113.4          \\
VQ-DVGO              & 31.77      & 0.954     & 1.4     &34.72      & 0.974     &1.3       & -      & -   & -     &   28.26    & 0.909     & 1.4          \\ \midrule \midrule
Plenoxels\cite{yu2021plenoxels}     & 31.71      & 0.958     & 259.8     &  34.12    &  0.977    &  283.3    & 26.43  & 0.842 & 2006.2 &  26.84     & 0.911     &  367.7      \\
VQ-Plenoxels         & 31.53      & 0.956     & 13.7     &  33.91    &  0.976    &  11.9    & 26.28  & 0.839 & 40.0  &  26.73     & 0.908     &  14.3       \\\midrule \midrule
TensoRF\cite{TensoRF}       & 33.09      & 0.963     & 67.6     &  36.72    &  0.982    &  71.6    & 26.70  & 0.836 & 179.8  & 28.54      & 0.921     & 72.6     \\
VQ-TensoRF           & 32.86      & 0.960     & 3.6      &  36.16    &  0.980    &  4.1     & 26.46  & 0.824 & 8.8    & 28.20      & 0.913     & 3.3      \\ \bottomrule
\end{tabular}
\caption{\textbf{Quantitative comparison.} We compare our VQRF with origin NeRF, uncompressed volumetric radiance fields and other methods focus on model size. Compared to all the baseline, our method achieve best psnr-size trade-off.}
\label{tab:big_table}
\end{table*}

\section{Experiments}
\label{sec:expperiments}

\subsection{Datasets}
\textbf{Synthetic-NeRF.}
The Synthetic-NeRF dataset was first introduced by \cite{NeRF}  and has been widely adopt by subsequent work. It contains 8 scenes rendered at 800$\times$800 by Blender. Each scene contains 100 rendered views as training set and 200 views for testing. 

\textbf{Synthetic-NSVF.}
The Synthetic-NSVF dataset \cite{SparseVoxel} includes eight objects with more complex geometry and lighting conditions. They are  rendered in the same resolution as Synthetic-NeRF.

\textbf{LLFF.}
 The LLFF dataset \cite{BenMildenhall2019LocalLF} was made of 8 real-world scenes captured by handheld mobile phone cameras. The views are forward-facing towards the scene. We use the images at 1008 $\times$ 756.

\textbf{Tanks \& Temples.}
The Tanks and Temples dataset \cite{Knapitsch2017} was reconstructed from video dataset which was captured in the real world. All the images are captured at 1920 $\times$ 1080 pixels. Here we use five scenarios \textit{(Barn, Caterpillar, Family, Ignatius, Truck)} which have no background.

\subsection{Implementation Detail}
When adapting VQRF to a specific method, we first obtain a non-compressed grid model following each method's default training config. The prune quantile $\beta_p$ is set to 0.001  for all datasets across all methods. The keep quantile $\beta_k$ is 0.6 for DVGO and Plenoxels, while 0.7 for TensoRF.
We chose 4096 as the default codebook size for all experiments. The codebook initialization takes 100 iterations with a batchsize of 10000 voxel points per iteration.
 Moving average factor $\lambda_d$ was set to 0.8, and ten dead codes was expired at every step. During the joint finetune phase, all learning rates are reverted to the midterm setting of their default training schedule. Please refer to the appendix for more details.

\textbf{VQ-DVGO.}
We use VQ-DVGO as the default method for all the experiments and ablation studies in section \ref{sec:expperiments}
, we joint finetune it for 10000 steps with 8196 rays per batch. The overall compressing time takes about 50$\%$ of the original training time across different scenes.

\textbf{VQ-Plenoxels.}
As Plenoxels have no neural network at all we only need to tune the vector quantized voxel grid in joint finetuning phase. We tune it with 25600 steps (equal to two epochs of original setting) with a batchsize of 5000, This compression pipeline brings about 20\% additional time cost upon original training time. 

\textbf{VQ-TensoRF.}
We choose TensoRF-VM-192 as the default model. As tensoRF utilize a triplane structure as the feature volume, we assign three different codebooks to the three planes for more extensive model capacity. Moreover,  three additional codebooks were used to apply vector quantization on density planes.

\begin{figure*}[t]  
    \centering
    \includegraphics[width=0.99\linewidth]{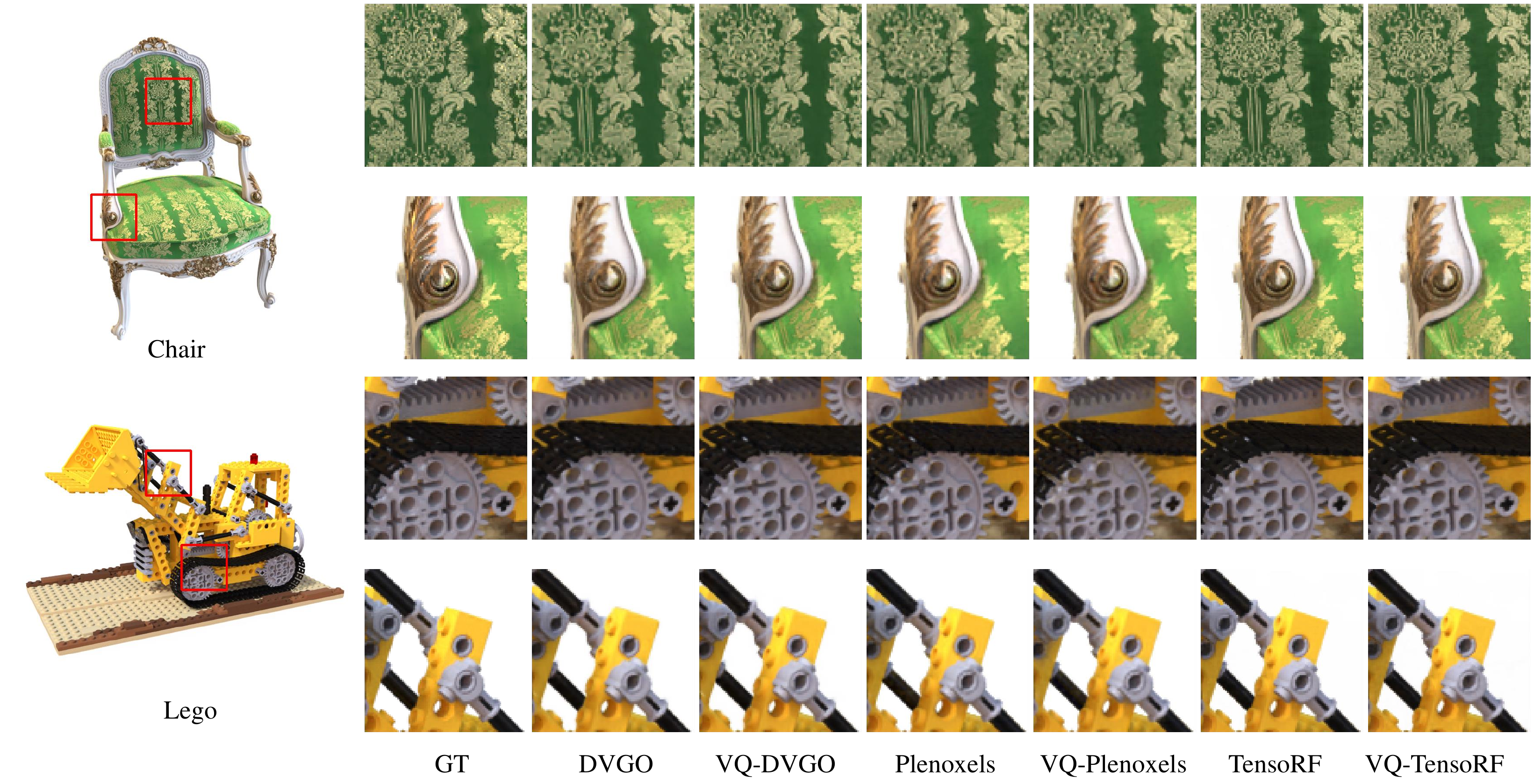}
    \caption{\textbf{Qualitative Comparison.} We can hardly observe visual artifacts on the rendering result of compressed model compared to its original model, even in the zoom-in images.} 
    \label{fig:nerf synthetic result} 
    \vspace{-0.5em}
\end{figure*}

\begin{figure*}[t]  
    \centering
    \includegraphics[width=0.9\linewidth]{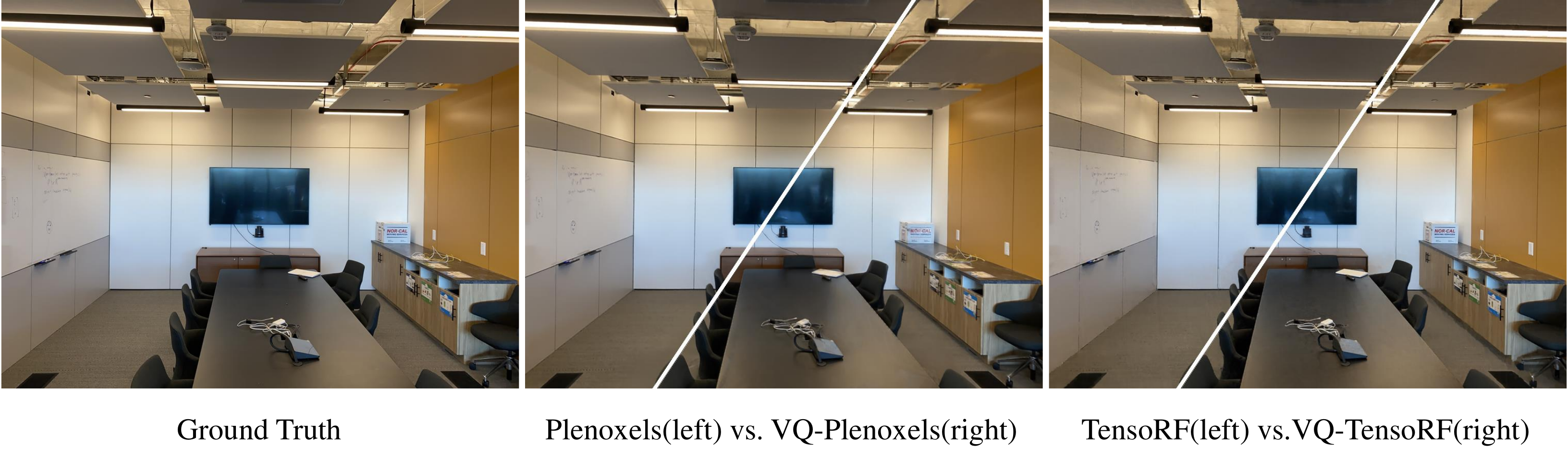}
    \caption{Visual quality comparison of origin model  versus our compressed model on real forward-facing dataset. } 
    \label{fig:llff result}   
    
\end{figure*}

\begin{figure} 
    \centering
    \includegraphics[width=\linewidth]{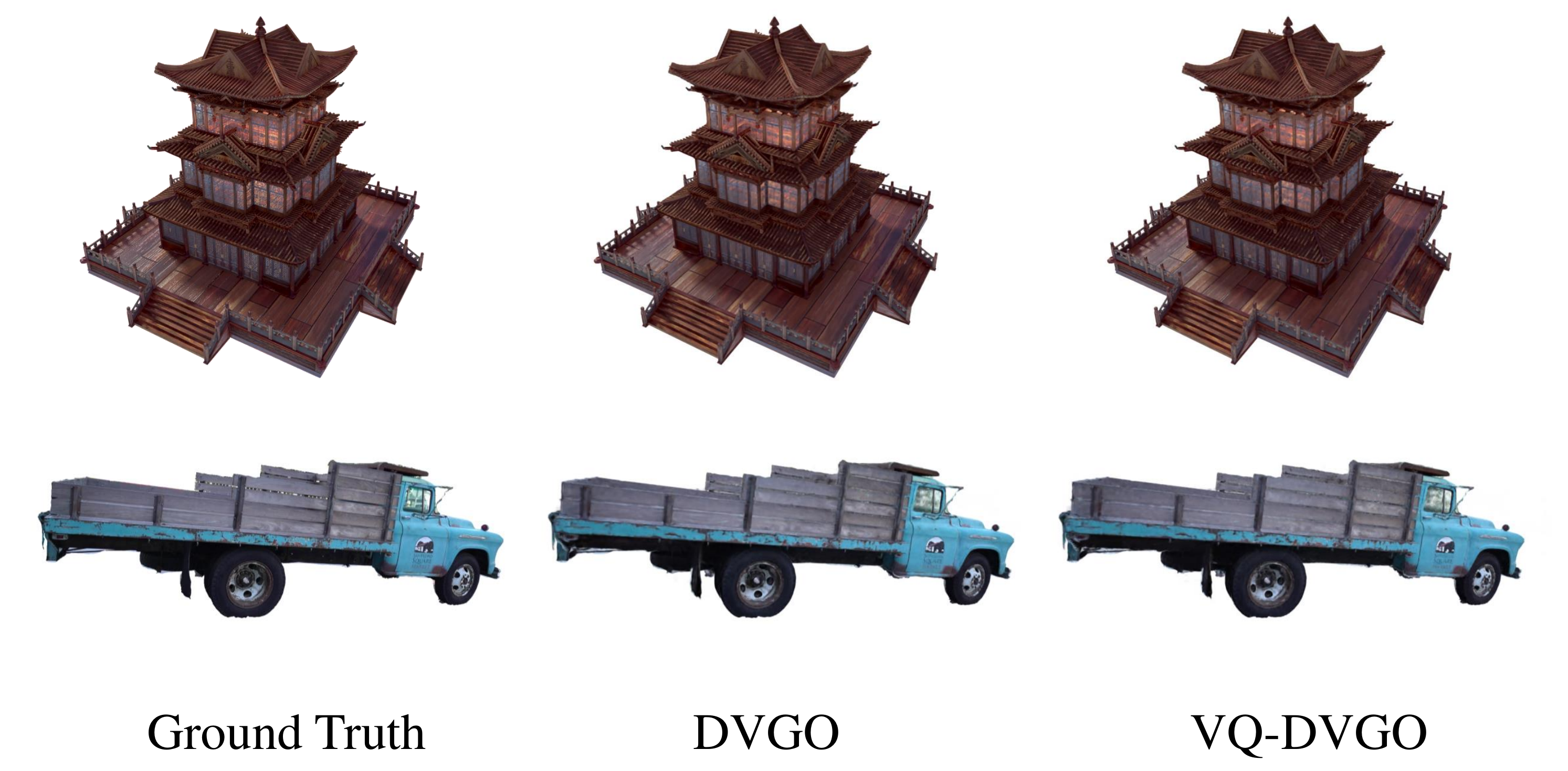}
    \caption{Rendering results on synthetic-NSVF dataset and Tanks\&Temples dataset show that our method can easily generalize to datasets with different distributions.} 
    \label{fig:nsvf result}   
\end{figure}

\subsection{Results}

\textbf{Quantitative result.}
We compare our work with original NeRF and other uncompressed volumetric radiance fields in Table \ref{tab:big_table}. Here the 'VQ-' prefix stands for "vector quantized",  i.e., the volumetric radiance fields compressed by our pipeline. All the reported model size of original DVGO, Plenoxels, and TensoRF was calculated after a standard zip compression for a fair comparison.

As shown in Table \ref{tab:big_table}, VQRF can realize satisfaction performance across all the combinations of methods and datasets. VQ-DVGO shows the achieved highest compression performance among the three methods,  realizing an average compression ratio of 75 $\times$ on the synthetic-NeRF dataset with negligible PSNR drop (0.13 dB). We believe that this performance advantage comes from the greater redundancy of DVGO compared to Plenoxels and TensoRF as illustrated in Figure \ref{fig:cdf_qq}. Nonetheless, despite the fact that Plenoxels already employed an empty voxel prune mechanism to reduce the model size, our pipeline still achieved a fairly good compression performance with over 20 $\times$ compression and a 0.2 dB performance drop in terms of PSNR. TensoRF has the smallest original model size, as it utilizes the decomposed tensor to represent the volumetric grid. This also means it has the lowest redundancy, but our VQ-TensoRF still enables a 20 $\times$ compression and has better performance in both model size and rendering quality even compared to its own compressed setting "TensoRF-CP".

\textbf{Visual Result.} We compare the rendering result of the compressed model and uncompressed model in figure \ref{fig:nerf synthetic result},\ref{fig:llff result},\ref{fig:nsvf result}, The visual difference is hard to be observed in both synthetic bounded scenes, forward-facing scenes, and real bounded scenes across all methods. 

\begin{figure}[t]  
    \centering
    \includegraphics[width=\linewidth]{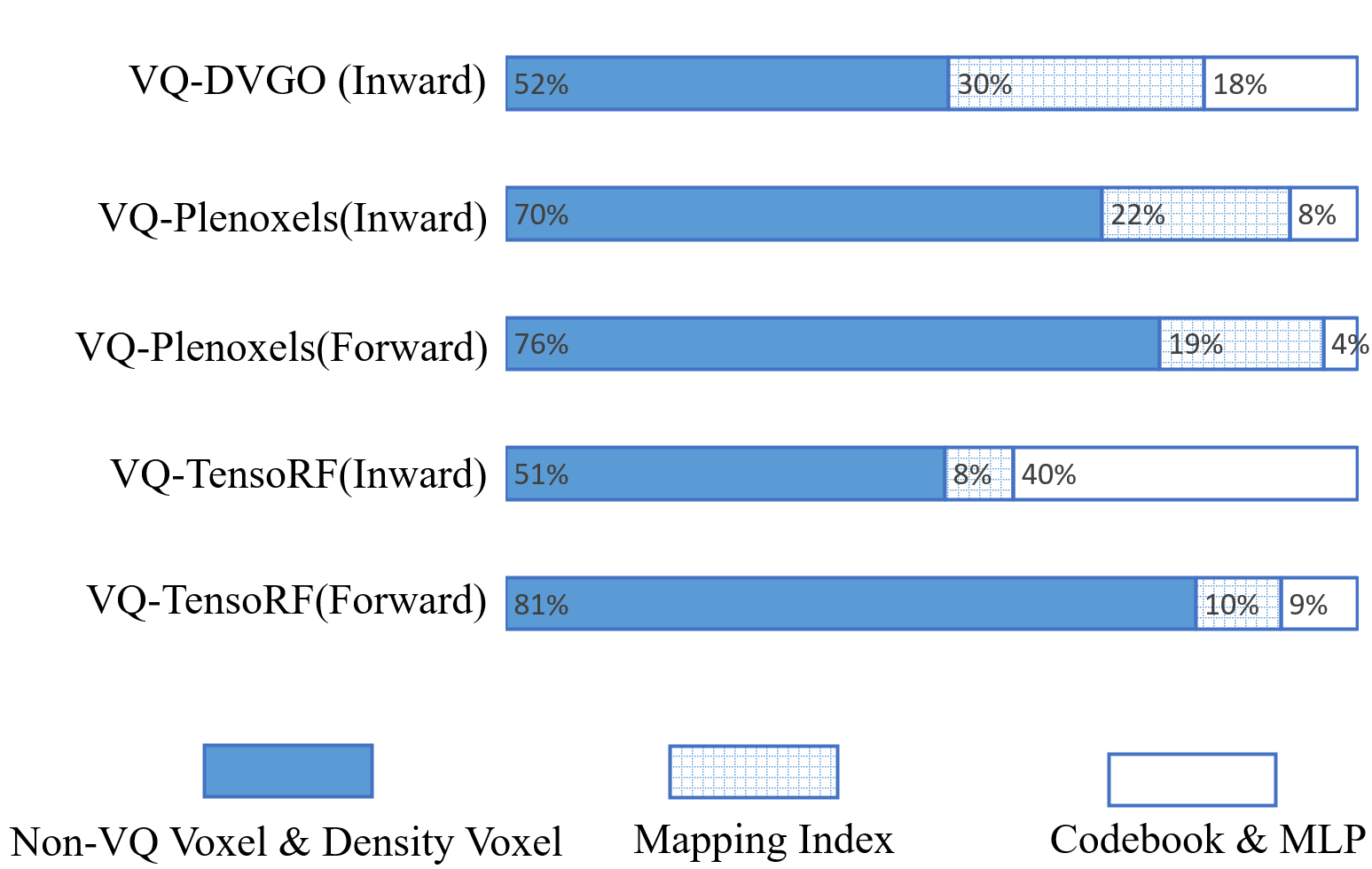}
    \caption{Visualization of model size proportion after compression. "Inward" represent the average results of synthetic-NeRF, synthetic-NSVF, and Tanks\&Temples datasets." Forward" represent results on LLFF datasets.} 
    \label{fig:storage_ratio}   
    \vspace{-0.5em}
\end{figure}

\textbf{Composition of final storage.}
Figure \ref{fig:storage_ratio} demonstrate the proportion of the three different part after applying our compression pipeline to volumetric radiance fields. Since metadata and MLP weights take a fairly small amount of storage, we combine their size with codebook's size for better visualization.  

\subsection{Ablation Study}
We conduct three ablation studies with VQ-DVGO on the synthetic-nerf dataset:

\noindent\textbf{Pruning and keeping percentage}.
\noindent We first compare different choices of  $\beta_p$ and  $\beta_k$. As shown in Table \ref{tab:Ablation of beta p and k}, smaller pruning parameter $\beta_p$ brings better rendering quality and enlarge model size in the meantime, while $\beta_k$ works in a contrary way, smaller choice degrades visual performance and improve compression ratio.

\noindent\textbf{Codebook size}.
Codebook size is another important hyperparameter that could be tuned, as shown in Table \ref{tab:ablation codebook size}, model size and rendering quality increase simultaneously as codebook size increases. A codebook larger than 4096 only brings minimal improvement on PSNR with an disproportionate additional storage.

\noindent\textbf{Step-by-step analysis}.
We conduct a step-by-step experiment to demonstrate the benefit of each module in the proposed framework, and the results are listed in Table \ref{tab:step by step}. We calculate the size after a zip compression for fair comparison. Compared to the uncompressed baseline, voxel pruning achieves 5$\times$ compression with a negligible PSNR drop, vector quantization bring another 5$\times$ reduction but decrease PSNR by 0.5dB. Joint finetuning recovers most of the rendering performance without affecting model size. Finally, our model size reaches 1 MB level by applying weight quantization.




\begin{table}[tbp]
    \centering
    \begin{tabular}{r|ccc|c}
        \toprule
         & PSNR$\uparrow$ & SSIM$\uparrow$ & LPIPS$\downarrow$ & SIZE$\downarrow$  \\
        \midrule
        16     &30.63  &0.944   &0.071     &1.040 \\
        64     &31.19  &0.949   &0.065     &1.134 \\
        256    &31.44  &0.951   &0.062     &1.164 \\
        1024   &31.62  &0.953   &0.059     &1.308 \\
        4096   &31.77  &0.954   &0.057     &1.431 \\
        16384  &31.81  &0.955   &0.056     &1.630 \\
        \bottomrule
    \end{tabular}
    \caption{
    {Ablation of codebook size.}
    }
    \label{tab:ablation codebook size}
    \vspace{-0.5em}
\end{table}

\begin{table}[tbp]
    \centering
    \small
    \begin{tabular}{l|ccc|c}
        \toprule
         & PSNR$\uparrow$ & SSIM$\uparrow$ & LPIPS$\downarrow$ & SIZE$\downarrow$  \\
        \midrule
  
        baseline              &31.90  &0.956   &0.054     &105.9 \\
        \midrule
        +voxel pruning          &31.88  &0.956   &0.054     &19.0 \\
        +vector quantization  &31.32  &0.952   &0.061    &4.8 \\
        +joint finetune        &31.79  &0.954   &0.036     &4.8 \\
        +weight quantization  &31.77  &0.954   &0.057     &1.4 \\
        \bottomrule
    \end{tabular}
    \caption{
    {Step-by-step analysis on performance gain.}
    }
    \label{tab:step by step}
    \vspace{-0.5em}
\end{table}

\begin{table}[tbp]
 \centering
\begin{tabular}{c|l|lll|l}
\toprule
\multicolumn{1}{c|}{$\beta_p$} & $\beta_k$ & PSNR$\uparrow$ & SSIM$\uparrow$ & LIPIS$\downarrow$ & SIZE$\downarrow$ \\ 
\midrule
\multirow{4}{*}{0}             & 0         &30.38 & 0.941 &0.077  &2.54      \\
                               & 0.3       &31.13 &0.948  &0.067  & 2.73    \\
                               & 0.6       &31.60 &0.953  &0.060  & 3.06     \\ 
                               & 0.9       &31.94 &0.956  &0.054 &3.86      \\
\midrule
\multirow{4}{*}{0.001$\dagger$}         & 0         & 31.04&0.947 &0.068  &0.93 \\
                               & 0.3       & 31.49&0.951 &0.062  &1.11 \\
                               & 0.6$\dagger$       & 31.77&0.954 &0.057  &1.43      \\ 
                               & 0.9       & 31.96&0.956 &0.054  &2.21     \\ 
\midrule
\multirow{4}{*}{0.01}          & 0         & 30.90&0.946 &0.070  & 0.74     \\
                               & 0.3       & 31.31&0.950 &0.040  & 0.92     \\
                               & 0.6       & 31.59&0.953 &0.059  & 1.24     \\ 
                               & 0.9       & 31.80&0.956 &0.055  &  2.02    \\
\midrule
\multirow{4}{*}{0.1}           & 0         &27.02  &0.920 &0.084  & 0.49      \\
                               & 0.3       &27.35  &0.935 & 0.077 & 0.67     \\
                               & 0.6       &27.64  &0.939 &0.054  & 0.99      \\
                               & 0.9       &28.27  &0.945 &0.065  & 1.67      \\

\bottomrule
\multicolumn{6}{l}{\scriptsize{$\dagger$ denote our default choice of  $\beta_p$ and $\beta_k$}}

\end{tabular}
\caption{Ablation of pruning quantile $\beta_p$ and keeping quantile $\beta_k$. }
\label{tab:Ablation of beta p and k}
\vspace{-0.5em}
\end{table}

\section{Conclusion}
In this paper we proposed VQRF,  a novel compression framework designed for volumetric radiance fields like DVGO and Plenoxels. Using an adaptive voxel pruning mechanism, a learnable vector quantization, and a simple weight quantization, we are able to compress the overall model size to 1 MB without degrading rendering quality. 
Extensive experiments demonstrate the effectiveness and generalization ability of VQRF, which achieves adequate performance on multiple methods across different datasets.

{\small
\bibliographystyle{ieee_fullname}
\bibliography{egbib}
}

\clearpage
\appendix


\twocolumn[
\centering
\Large
\textbf{Compressing Volumetric Radiance Fields to 1 MB} \\
\vspace{0.5em}Appendix \\
\vspace{1.0em}
] 
\appendix
\section{Detailed Results}

We present detailed results for each scene on the datasets used in the main paper, including synthetic-NeRF \cite{NeRF} in Table \ref{tab:syntheticnerf}, LLFF \cite{BenMildenhall2019LocalLF} in Table \ref{tab:llff},  synthetic-NSVF \cite{SparseVoxel} in  Table \ref{tab:nsvf} and Tanks\&Temples \cite{Knapitsch2017} in Table  \ref{tab:tnt}, respectively. The detailed comparison validate that our method can achieve comparable rendering quality with significant advantage on model size overhead compared to original DVGO\cite{ChengSun2022DirectVG}, Plenoxels\cite{yu2021plenoxels} and TensoRF\cite{TensoRF}, demonstrating the effectiveness and generalizability of our method.

 \section{More Qualitative Comparisons}
\begin{figure*}
    \centering
    \includegraphics[width=0.7\linewidth]{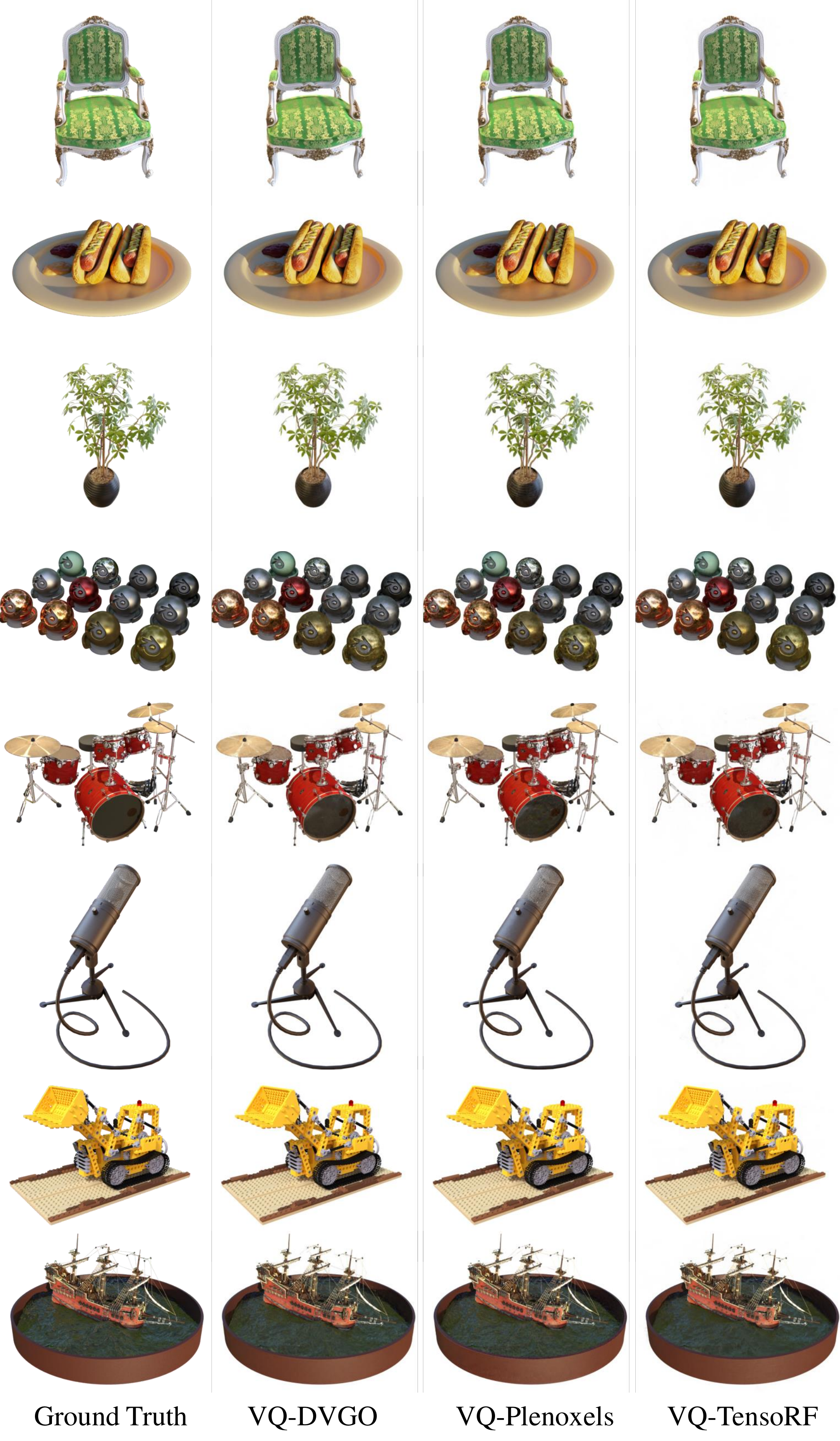}
    \caption{\textbf{NeRF-Synthetic scenes.} We show a random view for each scene in the dataset, comparing ground truth with our VQ-DVGO, VQ-Plenoxels, VQ-TensoRF.}
    \label{fig:syn-result}
\end{figure*}

\begin{figure*}
    \centering
    \includegraphics[width=0.58\linewidth]{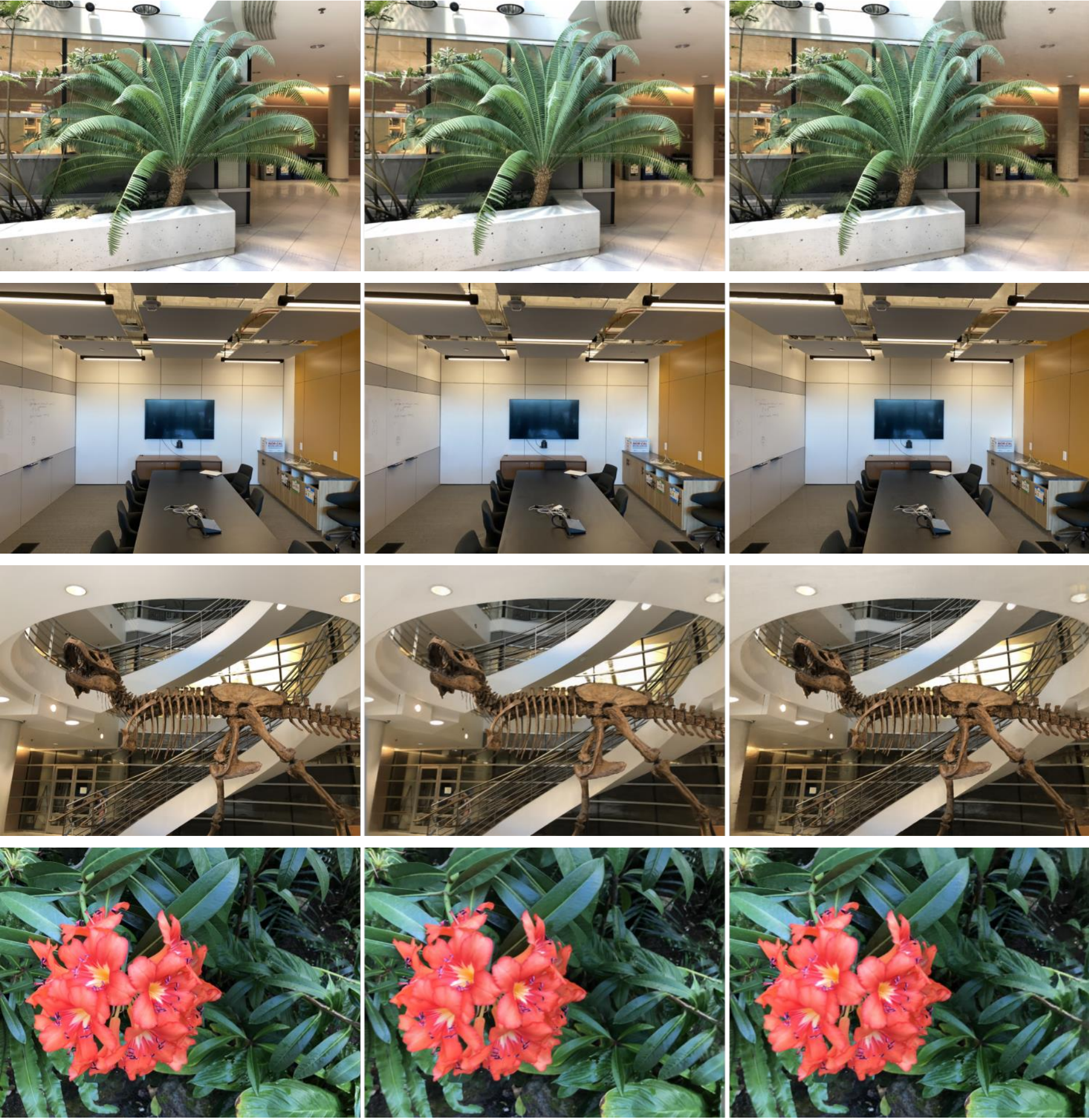}
    \includegraphics[width=0.58\linewidth]{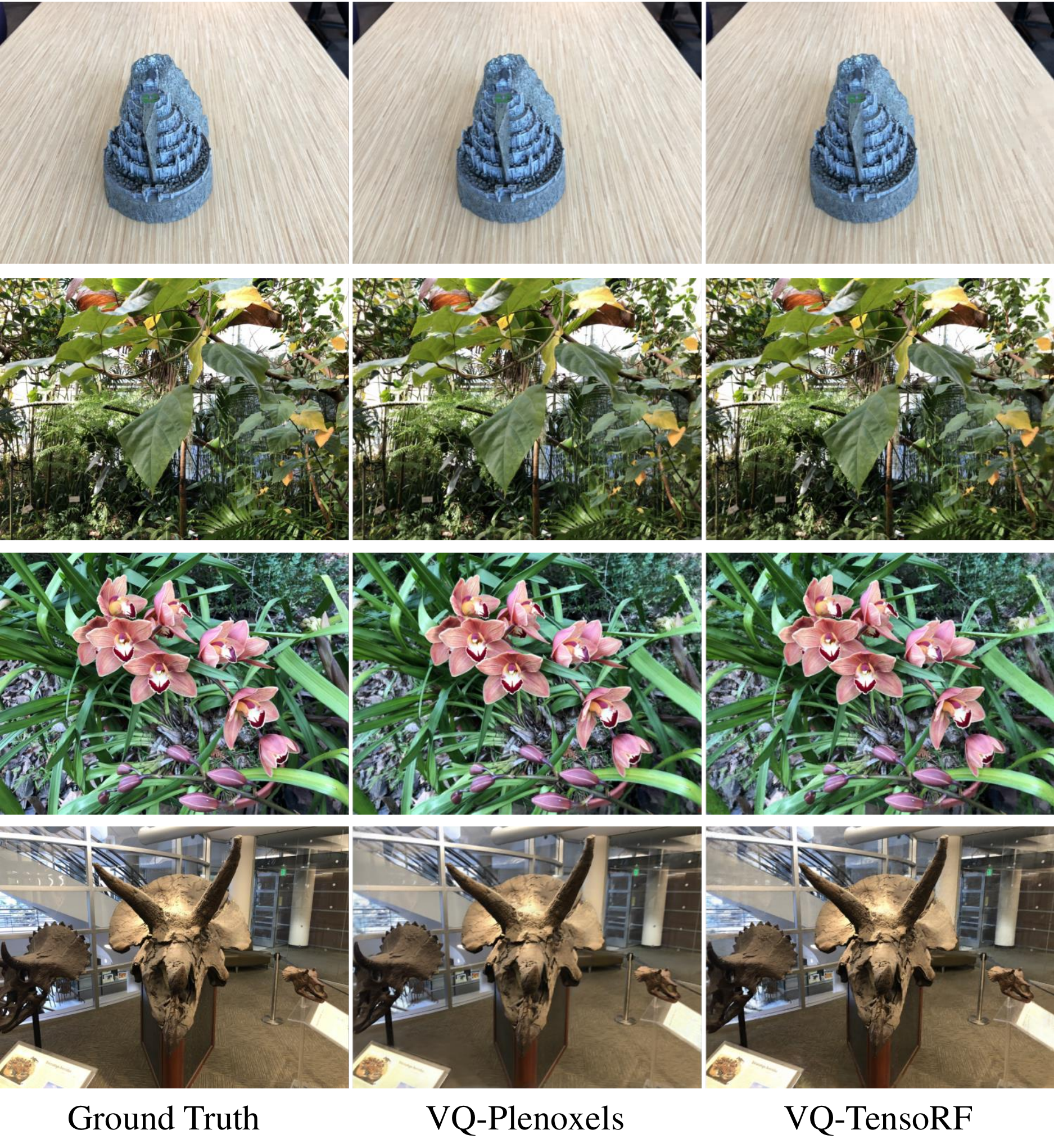}
    \caption{\textbf{LLFF scenes.} We show a random view for each scene in the dataset, comparing ground truth with our VQ-Plenoxels, VQ-TensoRF.}
    \label{fig:llff-result}
\end{figure*}

\begin{figure*}
    \centering
    \includegraphics[width=0.7\linewidth]{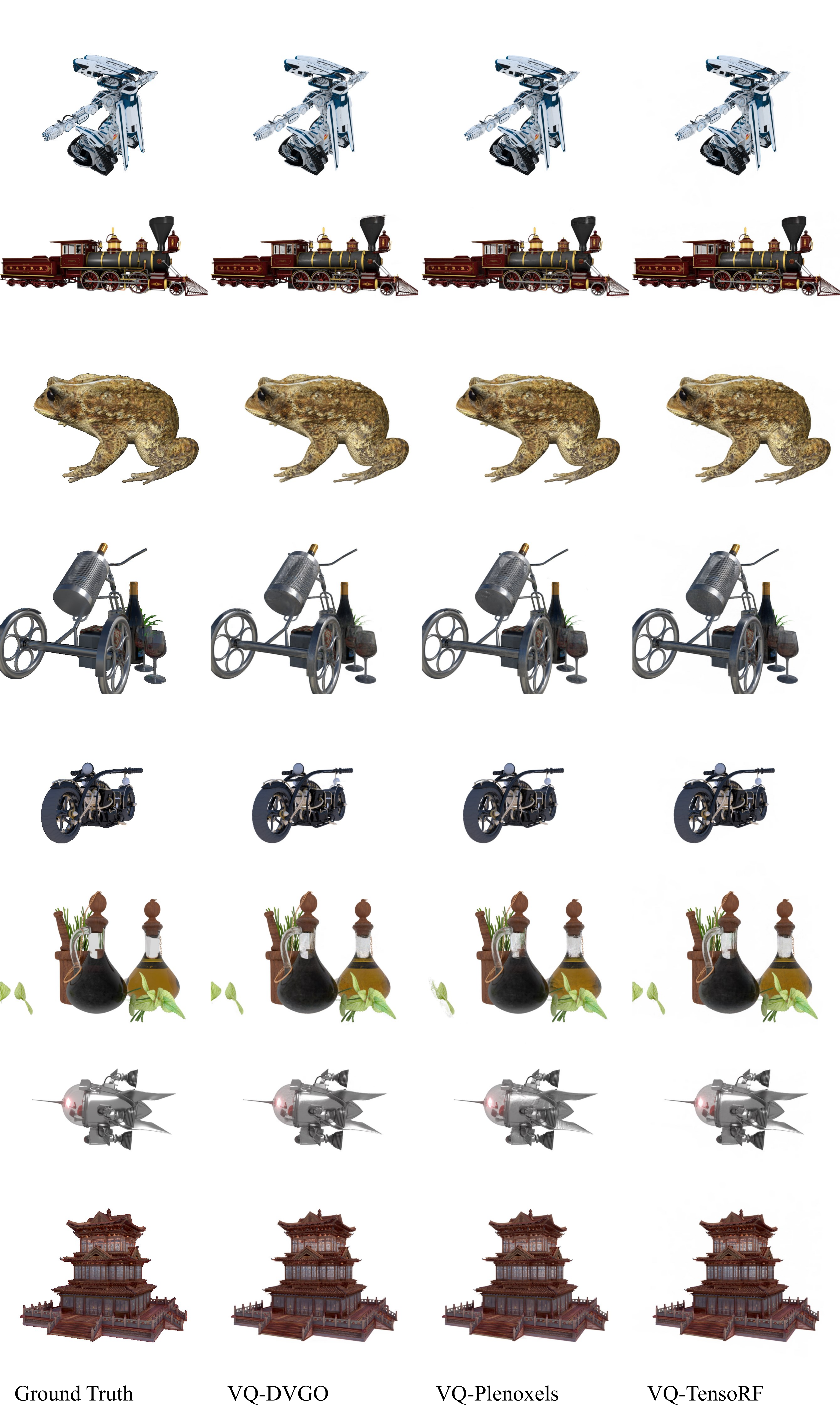}
    \caption{\textbf{Synthetic-NSVF scenes.} We show a random view for each scene in the dataset, comparing ground truth with our VQ-DVGO, VQ-Plenoxels, VQ-TensoRF.}
    \label{fig:nsvf-result}
\end{figure*}

\begin{figure*}
    \centering
    \includegraphics[width=0.9\linewidth]{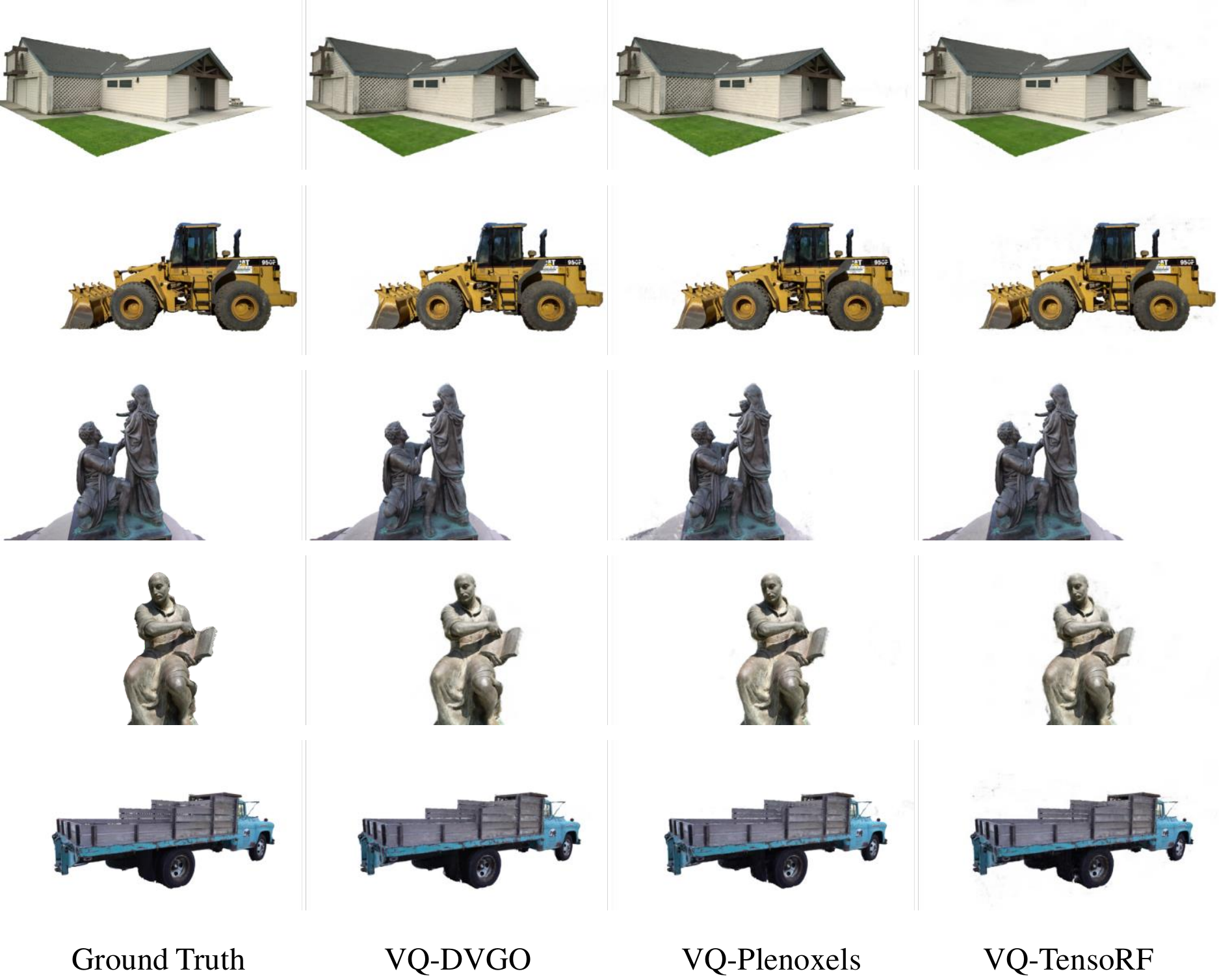}
    \caption{\textbf{Tanks\&Temples scenes.} We show a random view for each scene in the dataset, comparing ground truth with our VQ-DVGO, VQ-Plenoxels, VQ-TensoRF.}
    \label{fig:tat-result}
\end{figure*}

We further show rendering a randomly selected view of example scenes on the datasets in Figure \ref{fig:syn-result},  \ref{fig:llff-result},  \ref{fig:nsvf-result} and \ref{fig:tat-result}, respectively. 
The rendering quality of the model after compression via our method can be effectively preserved compared to the model before compression, even on fine details, meanwhile we can achieve ~$100\times$ lower storage cost via our compression framework.

\section{Implementation Details}
The model of DVGO consists of density grid, feature grid and a shallow MLP for color estimation. As mentioned in the main paper, we first prune off less important voxels according to the quantile of the cumulative score rate.
We only use the voxels that need to be vector quantized for codebook initialization. 
After codebook initialization, we start joint finetuning with the learning rate 8e-2 for density grid, 1.6e-3 for both the feature grid and the shallow MLP.  We adopt a exponential learning rate decay schedule following its original setup, downscaling the learning rates by 0.3 for every 10k iterations. The finetuning stage takes 10000 iterations with 8192 rays per iteration. 

Plenoxels consist of a density grid and a feature grid filled with spherical harmonic coefficients. During joint finetune stage. The fintune stage takes 25600 iterations, which are equal to two epochs in the original Plenoxels, by resetting the learning rate scheduler to the sixth epoch in the origin setting.

For TensoRF, we leverage a virtual grid  and compute importance scores for all the grid points. Then the points are projected onto tri-planes by aggregating them along each axis. The finetuneing stage for TensoRF takes 10000 iterations, with the learning rate 5e-3 for all the density tri-plane, feature tri-plane and the shallow MLP. We adopt an exponential learning rate decay for TensoRF, downscaling the learning rate by 0.3 for every 30k iterations. 








\begin{table*}
 
	\renewcommand{\arraystretch}{1.2}
	\centering
    \resizebox{\textwidth}{!}{
		\begin{tabular}{c c c c c c c c c c c c c c c c c c c }
        \toprule
        &\multicolumn{9}{c}{\bf Synthetic-NeRF}\\
        &Method & Chair & Drums & Ficus & Hotdog & Lego & Materials & Mic & Ship &\bf Avg.\\
         \hline
        \hline
         \multirow{6}{*}{Size(MB) ($\downarrow$)}                  
            & DVGO  & 99.86  & 90.64  & 103.98 & 124.73 & 118.56 & 163.77 & 47.03 & 98.79  & 105.92 \\
            & VQ-DVGO & 0.99   & 0.89   & 1.06   & 1.46   & 1.48   & 3.11   & 0.41  & 2.05   & 1.43   \\
         \cline{2-11}
            & Plenoxels & 186.17 & 161.51 & 108.46 & 292.43 & 294.53 & 196.51 & 82.27 & 756.37 & 259.78 \\
            & VQ-Plenoxels & 10.12  & 7.17   & 4.31   & 17.45  & 13.56  & 4.09   & 12.35 & 40.41  & 13.68  \\
            
            \cline{2-11}
            & TensoRF   & 62.99  & 63.20  & 66.00  & 78.06  & 63.88  & 78.51  & 62.09 & 65.74  & 67.56  \\
            & VQ-TensoRF & 3.47   & 3.31   & 4.12   & 3.52   & 3.57   & 4.27   & 2.44  & 3.76   & 3.56   \\
         \hline
        \hline
         \multirow{6}{*}{PSNR(dB) ($\uparrow $)} 
            & DVGO    & 34.09  & 25.47  & 32.66  & 36.67  & 34.59  & 29.51  & 33.11 & 29.11  & 31.90  \\
            & VQ-DVGO & 33.80  & 25.38  & 32.67  & 36.47  & 34.27  & 29.28  & 33.11 & 29.24  & 31.77  \\
            \cline{2-11}
            & Plenoxels   & 33.99  & 25.35  & 31.83  & 36.42  & 34.10  & 29.14  & 33.27 & 29.62  & 31.71  \\
            &VQ-Plenoxels & 33.82  & 25.30  & 31.87  & 36.01  & 33.66  & 28.89  & 33.24 & 29.45  & 31.53  \\
            \cline{2-11}
            & TensoRF     & 35.61  & 25.98  & 33.95  & 37.40  & 36.36  & 30.03  & 34.82 & 30.57  & 33.09  \\
            & VQ-TensoRF & 35.10  & 25.97  & 33.85  & 36.98  & 36.03  & 30.07  & 34.45 & 30.38  & 32.86  \\
         \hline
         \hline 
         \multirow{6}{*}{SSIM($\uparrow$)}
            & DVGO       & 0.976  & 0.930  & 0.978  & 0.980  & 0.976  & 0.950  & 0.983 & 0.878  & 0.956  \\
            & VQ-DVGO & 0.974  & 0.928  & 0.977  & 0.978  & 0.973  & 0.945  & 0.982 & 0.877  & 0.954  \\
            \cline{2-11}
            & Plenoxels  & 0.977  & 0.933  & 0.976  & 0.980  & 0.975  & 0.949  & 0.985 & 0.890  & 0.958  \\
            &VQ-Plenoxels & 0.975  & 0.931  & 0.975  & 0.979  & 0.972  & 0.945  & 0.984 & 0.889  & 0.956  \\
            \cline{2-11}
            & TensoRF     & 0.984  & 0.937  & 0.982  & 0.982  & 0.983  & 0.952  & 0.988 & 0.892  & 0.963  \\
            & VQ-TensoRF & 0.981  & 0.932  & 0.982  & 0.980  & 0.981  & 0.950  & 0.986 & 0.887  & 0.960  \\
         \hline
        \hline
         \multirow{6}{*}{LPIPS$_{ALEX}$($\downarrow$)}
            & DVGO       & 0.017  & 0.060  & 0.015  & 0.018  & 0.013  & 0.027  & 0.014 & 0.117  & 0.035  \\
            & VQ-DVGO & 0.018  & 0.061  & 0.017  & 0.018  & 0.013  & 0.033  & 0.014 & 0.112  & 0.036  \\
            \cline{2-11}
            & Plenoxels  & 0.019  & 0.055  & 0.015  & 0.018  & 0.016  & 0.012  & 0.026 & 0.083  & 0.031  \\
            & VQ-Plenoxels & 0.020  & 0.057  & 0.016  & 0.021  & 0.018  & 0.031  & 0.013 & 0.089  & 0.033  \\
            \cline{2-11}
            & TensoRF    & 0.010  & 0.051  & 0.013  & 0.014  & 0.007  & 0.027  & 0.008 & 0.087  & 0.027  \\
            & VQ-TensoRF & 0.016  & 0.063  & 0.014  & 0.017  & 0.009  & 0.030  & 0.013 & 0.094  & 0.032  \\
        \hline
        \hline
        \multirow{6}{*}{LPIPS$_{VGG}$($\downarrow$)}
            & DVGO        & 0.028  & 0.078  & 0.025  & 0.034  & 0.027  & 0.059  & 0.018 & 0.160  & 0.054  \\
            &VQ-DVGO & 0.032  & 0.082  & 0.028  & 0.039  & 0.030  & 0.066  & 0.020 & 0.160  & 0.057  \\
            \cline{2-11}
            & Plenoxels   & 0.031  & 0.067  & 0.026  & 0.038  & 0.028  & 0.057  & 0.015 & 0.134  & 0.050  \\
            & VQ-Plenoxels & 0.033  & 0.072  & 0.028  & 0.041  & 0.033  & 0.064  & 0.017 & 0.139  & 0.053  \\
            \cline{2-11}
            & TensoRF     & 0.022  & 0.072  & 0.023  & 0.032  & 0.018  & 0.060  & 0.015 & 0.141  & 0.048  \\
            & VQ-TensoRF & 0.035  & 0.099  & 0.028  & 0.040  & 0.024  & 0.064  & 0.025 & 0.149  & 0.058    \\
         \bottomrule
    
	\end{tabular}
    }
\caption{Per-scene results on Synthetic-NeRF \cite{NeRF}.}
\label{tab:syntheticnerf}   
\end{table*}

\begin{table*}[]

	\renewcommand{\arraystretch}{1.2}
	\centering
    \resizebox{\textwidth}{!}{
		\begin{tabular}{c c c c c c c c c c c c c c c c c c c }
  
        \toprule
        &\multicolumn{9}{c}{\bf LLFF}\\
            &  Method     & Fern    & Flower  & Room    & Leaves  & Horns   & Trex    & Fortress & Orchids & \bf Avg.    \\ 
            \hline\hline
            \multirow{4}{*}{Size(MB) ($\downarrow$)}      
            & Plenoxels & 1842.22 & 1719.30 & 1613.74 & 2061.28 & 1959.84 & 1819.81 & 1685.97  & 3347.22 & 2006.17 \\
            & VQ-Plenoxels & 38.94   & 38.92   & 31.40   & 41.71   & 38.34   & 34.64   & 38.82    & 57.40   & 40.02   \\\cline{2-11}
            & TensoRF  & 179.92  & 179.81  & 179.87  & 179.70  & 179.81  & 179.85  & 179.87   & 179.90  & 179.84  \\
            & VQ-TensoRF & 8.61    & 8.97    & 8.02    & 9.14    & 8.40    & 8.11    & 9.32     & 9.20    & 8.72    \\
            \hline
            \hline
            \multirow{4}{*}{PSNR(dB) ($\uparrow$)}
            & Plenoxels & 25.51   & 28.16   & 30.29   & 21.58   & 27.68   & 26.51   & 31.10    & 20.65   & 26.44   \\
            & VQ-Plenoxels & 25.46   & 27.91   & 30.18   & 21.50   & 27.52   & 26.11   & 30.93    & 20.53   & 26.27   \\\cline{2-11}
            & TensoRF  & 25.03   & 28.10   & 32.16   & 21.12   & 28.31   & 27.56   & 31.44    & 19.85   & 26.70   \\
            & VQ-TensoRF & 24.82   & 27.82   & 31.89   & 21.00   & 27.96   & 27.30   & 31.14    & 19.75   & 26.46   \\
            \hline
            \hline
            \multirow{4}{*}{SSIM($\uparrow$)}
            & Plenoxels & 0.835   & 0.866   & 0.938   & 0.764   & 0.859   & 0.891   & 0.886    & 0.698   & 0.842   \\
            & VQ-Plenoxels & 0.833   & 0.861   & 0.936   & 0.761   & 0.856   & 0.888   & 0.884    & 0.691   & 0.839   \\\cline{2-11}
            & TensoRF  & 0.801   & 0.857   & 0.952   & 0.744   & 0.883   & 0.910   & 0.898    & 0.644   & 0.836   \\
            & VQ-TensoRF & 0.791   & 0.843   & 0.947   & 0.727   & 0.866   & 0.902   & 0.881    & 0.636   & 0.824   \\
            \hline
            \hline
            \multirow{4}{*}{LPIPS$_{ALEX}$($\downarrow$)}
            & Plenoxels & 0.150   & 0.122   & 0.128   & 0.153   & 0.178   & 0.132   & 0.108    & 0.187   & 0.145   \\
            & VQ-Plenoxels & 0.146   & 0.119   & 0.126   & 0.146   & 0.176   & 0.131   & 0.105    & 0.184   & 0.142   \\\cline{2-11}
            & TensoRF  & 0.157   & 0.103   & 0.076   & 0.144   & 0.103   & 0.080   & 0.067    & 0.192   & 0.115   \\
            & VQ-TensoRF & 0.166   & 0.115   & 0.084   & 0.156   & 0.125   & 0.089   & 0.098    & 0.202   & 0.129   \\
            \hline
        \hline
        \multirow{4}{*}{LPIPS$_{VGG}$($\downarrow$)}
            & Plenoxels & 0.835   & 0.866   & 0.938   & 0.764   & 0.859   & 0.891   & 0.886    & 0.698   & 0.842   \\
            & VQ-Plenoxels & 0.220   & 0.180   & 0.199   & 0.200   & 0.232   & 0.238   & 0.178    & 0.241   & 0.211   \\\cline{2-11}
            & TensoRF  & 0.249   & 0.178   & 0.162   & 0.221   & 0.182   & 0.201   & 0.143    & 0.281   & 0.202   \\
            & VQ-TensoRF & 0.263   & 0.200   & 0.173   & 0.249   & 0.214   & 0.218   & 0.183    & 0.294   & 0.224  \\
        \bottomrule
        \end{tabular}
}
\caption{Per-scene Results on  LLFF \cite{BenMildenhall2019LocalLF}.}
\label{tab:llff}
\end{table*}

\begin{table*}[]
	\renewcommand{\arraystretch}{1.2}
	\centering
    \resizebox{\textwidth}{!}{
		\begin{tabular}{c c c c c c c c c c c c c c c c c c c }
        \toprule
        &\multicolumn{9}{c}{\bf Synthetic-NSVF}\\
            &     Method   & Bike   & Lifestyle & Palace & Robot  & Spaceship & Steamtrain & Toad   & Wineholder & \bf Avg.   \\
            \hline\hline
            \multirow{6}{*}{Size(MB) ($\downarrow$)}      
            & DVGO  & 114.91 & 103.68    & 109.25 & 102.15 & 132.30    & 156.25     & 133.37 & 106.40     & 119.79 \\
            & VQ-DVGO & 1.08   & 1.08      & 1.50   & 0.99   & 1.67      & 1.98       & 0.85   & 0.94       & 1.26   \\\cline{2-11}
            & Plenoxels & 89.25  & 361.27    & 629.13 & 137.08 & 183.12    & 91.79      & 610.44 & 164.62     & 283.34 \\
            & VQ-Plenoxels & 4.14   & 15.88     & 27.54  & 6.21   & 6.80      & 4.17       & 22.72  & 7.57       & 11.88  \\\cline{2-11}
            & TensoRF  & 73.53  & 67.56     & 67.37  & 70.73  & 70.60     & 83.71      & 71.44  & 68.18      & 71.64  \\
            & VQ-TensoRF & 3.75   & 4.47      & 3.70   & 3.69   & 4.24      & 4.49       & 5.18   & 3.88       & 4.17   \\
            \hline
            \hline
            \multirow{6}{*}{PSNR(dB) ($\uparrow$)}
            & DVGO  & 38.14  & 33.74     & 34.46  & 36.38  & 37.53     & 35.43      & 32.99  & 30.26      & 34.87  \\
            & VQ-DVGO & 37.89  & 33.65     & 34.42  & 36.06  & 37.51     & 35.32      & 32.68  & 30.25      & 34.72  \\\cline{2-11}
            & Plenoxels & 37.83  & 31.04     & 35.30  & 35.91  & 34.36     & 34.21      & 34.34  & 30.01      & 34.12  \\
            & VQ-Plenoxels & 37.47  & 30.93     & 34.94  & 35.61  & 34.25     & 34.00      & 34.16  & 29.91      & 33.91  \\\cline{2-11}
            & TensoRF  & 39.39  & 34.64     & 37.84  & 38.55  & 38.74     & 37.99      & 35.10  & 31.49      & 36.72  \\
            & VQ-TensoRF & 38.67  & 34.46     & 37.42  & 37.95  & 38.36     & 37.60      & 33.54  & 31.36      & 36.17  \\
            \hline
            \hline
            \multirow{6}{*}{SSIM($\uparrow$)}
            & DVGO  & 0.991  & 0.965     & 0.962  & 0.992  & 0.987     & 0.987      & 0.965  & 0.950      & 0.975  \\
            & VQ-DVGO & 0.991  & 0.964     & 0.961  & 0.991  & 0.987     & 0.987      & 0.963  & 0.950      & 0.974  \\\cline{2-11}
            & Plenoxels & 0.992  & 0.967     & 0.974  & 0.991  & 0.981     & 0.983      & 0.976  & 0.959      & 0.978  \\
            & VQ-Plenoxels & 0.991  & 0.965     & 0.972  & 0.990  & 0.981     & 0.981      & 0.974  & 0.957      & 0.976  \\\cline{2-11}
            & TensoRF  & 0.993  & 0.969     & 0.981  & 0.995  & 0.989     & 0.991      & 0.979  & 0.962      & 0.982  \\
            & VQ-TensoRF & 0.992  & 0.967     & 0.978  & 0.994  & 0.988     & 0.989      & 0.970  & 0.960      & 0.980  \\
            \hline
            \hline
            \multirow{6}{*}{LPIPS$_{ALEX}$($\downarrow$)}
            & DVGO  & 0.004  & 0.026     & 0.027  & 0.005  & 0.009     & 0.011      & 0.029  & 0.036      & 0.018  \\
            & VQ-DVGO & 0.004  & 0.026     & 0.025  & 0.005  & 0.010     & 0.011      & 0.029  & 0.035      & 0.018  \\\cline{2-11}
             & Plenoxels & 0.004  & 0.030     & 0.016  & 0.006  & 0.017     & 0.017      & 0.019  & 0.027      & 0.017  \\
            & VQ-Plenoxels & 0.005  & 0.032     & 0.018  & 0.006  & 0.018     & 0.018      & 0.021  & 0.029      & 0.018  \\\cline{2-11}
            & TensoRF  & 0.003  & 0.020     & 0.010  & 0.003  & 0.009     & 0.006      & 0.014  & 0.022      & 0.011  \\
            & VQ-TensoRF & 0.003  & 0.022     & 0.011  & 0.003  & 0.010     & 0.007      & 0.022  & 0.024      & 0.013  \\
            \hline
        \hline
        \multirow{6}{*}{LPIPS$_{VGG}$($\downarrow$)}
            & DVGO  & 0.011  & 0.053     & 0.043  & 0.013  & 0.020     & 0.022      & 0.046  & 0.054      & 0.033  \\
            & VQ-DVGO & 0.013  & 0.056     & 0.043  & 0.013  & 0.022     & 0.023      & 0.047  & 0.056      & 0.034  \\\cline{2-11}
              & Plenoxels & 0.011  & 0.047     & 0.026  & 0.013  & 0.025     & 0.030      & 0.031  & 0.046      & 0.029  \\
            & VQ-Plenoxels & 0.013  & 0.050     & 0.029  & 0.014  & 0.026     & 0.033      & 0.032  & 0.050      & 0.031  \\\cline{2-11}
            & TensoRF  & 0.010  & 0.046     & 0.021  & 0.010  & 0.020     & 0.017      & 0.028  & 0.048      & 0.025  \\
            & VQ-TensoRF & 0.013  & 0.051     & 0.024  & 0.011  & 0.022     & 0.023      & 0.044  & 0.053      & 0.030  \\
            \bottomrule
\end{tabular}
}
\caption{Per-scene Results on Synthetic-NSVF\cite{SparseVoxel}.}
\label{tab:nsvf}
\end{table*}

\begin{table*}[]
	\renewcommand{\arraystretch}{1.2}
	\centering
    \resizebox{0.8\textwidth}{!}{
		\begin{tabular}{c c c c c c c c c c c c }
        \toprule
        &\multicolumn{6}{c}{\bf Tanks\&Temples}\\
            & Method  & Barn   & Caterpillar & Family & Ignatius & Truck  &\bf Avg.    \\
            \hline\hline
            \multirow{6}{*}{Size(MB) ($\downarrow$)}      
            & DVGO       & 137.69 & 116.77      & 97.71  & 102.08   & 112.73 & 113.40 \\
            & VQ-DVGO & 1.82   & 1.46        & 1.20   & 1.12     & 1.42   & 1.40   \\\cline{2-8}
            & Plenoxels & 373.69 & 337.52      & 527.12 & 341.43   & 258.76 & 367.71 \\
            & VQ-Plenoxels & 11.97  & 13.40       & 20.90  & 15.63    & 9.43   & 14.27  \\\cline{2-8}
            & TensoRF  & 80.82  & 72.07       & 67.11  & 67.23    & 75.95  & 72.64  \\
            & VQ-TensoRF & 3.04   & 3.47        & 2.79   & 3.46     & 3.51   & 3.25   \\
            \hline
            \hline
            \multirow{6}{*}{PSNR(dB) ($\uparrow$)}   
            & DVGO     & 26.80  & 25.67       & 33.74  & 28.20    & 27.08  & 28.30  \\
            &   VQ-DVGO & 26.76  & 25.66       & 33.66  & 28.23    & 27.00  & 28.26  \\\cline{2-8}
            & Plenoxels& 24.57  & 25.18       & 30.03  & 27.86    & 26.55  & 26.84  \\
            & VQ-Plenoxels & 24.53  & 24.99       & 29.93  & 27.76    & 26.43  & 26.73  \\\cline{2-8}
            & TensoRF  & 27.48  & 25.92       & 34.06  & 28.38    & 26.89  & 28.54  \\
            & VQ-TensoRF & 27.11  & 25.59       & 33.43  & 28.27    & 26.59  & 28.20  \\
            \hline\hline
            \multirow{6}{*}{SSIM ($\uparrow$)}      
            & DVGO  & 0.837  & 0.903       & 0.962  & 0.943    & 0.905  & 0.910  \\
            & VQ-DVGO & 0.837  & 0.901       & 0.961  & 0.942    & 0.903  & 0.909  \\\cline{2-8}
            & Plenoxels & 0.842  & 0.904       & 0.959  & 0.942    & 0.909  & 0.911  \\
            & VQ-Plenoxels & 0.839  & 0.899       & 0.957  & 0.939    & 0.906  & 0.908  \\\cline{2-8}
            & TensoRF  & 0.866  & 0.910       & 0.966  & 0.949    & 0.913  & 0.921  \\
            & VQ-TensoRF & 0.857  & 0.902       & 0.960  & 0.944    & 0.903  & 0.913  \\
             \hline
            \hline
            \multirow{6}{*}{LPIPS$_{ALEX}$($\downarrow$)}    
            & DVGO  & 0.292  & 0.152       & 0.063  & 0.092    & 0.146  & 0.149  \\
            & VQ-DVGO & 0.287  & 0.150       & 0.060  & 0.089    & 0.143  & 0.146  \\\cline{2-8}
            & Plenoxels & 0.277  & 0.164       & 0.075  & 0.094    & 0.152  & 0.153  \\
            & VQ-Plenoxels & 0.291  & 0.162       & 0.075  & 0.102    & 0.152  & 0.156  \\\cline{2-8}
            & TensoRF  & 0.208  & 0.135       & 0.053  & 0.076    & 0.126  & 0.120  \\
            & VQ-TensoRF & 0.231  & 0.167       & 0.063  & 0.084    & 0.155  & 0.140  \\
            \hline\hline
            \multirow{6}{*}{LPIPS$_{VGG}$ ($\downarrow$)}      
            & DVGO  & 0.294  & 0.170       & 0.070  & 0.087    & 0.161  & 0.156  \\
            & VQ-DVGO & 0.296  & 0.173       & 0.070  & 0.087    & 0.163  & 0.158  \\\cline{2-8}
            & Plenoxels & 0.277  & 0.164       & 0.075  & 0.094    & 0.152  & 0.153  \\
            & VQ-Plenoxels & 0.284  & 0.179       & 0.080  & 0.103    & 0.160  & 0.161  \\\cline{2-8}
            & TensoRF  & 0.248  & 0.160       & 0.060  & 0.077    & 0.148  & 0.139  \\
            & VQ-TensoRF & 0.275  & 0.193       & 0.075  & 0.088    & 0.182  & 0.163  \\
            \bottomrule
\end{tabular}
}
\caption{Per-scene Results on Tanks\&Temples\cite{Knapitsch2017}.}
\label{tab:tnt}
\end{table*}



\end{document}